\documentclass[10pt,twocolumn,letterpaper]{article}

\usepackage[pagenumbers]{cvpr} 

\usepackage{graphicx}
\usepackage{amsmath}
\usepackage{amssymb}
\usepackage{booktabs}

\usepackage[pagebackref,breaklinks,colorlinks]{hyperref}

\usepackage[capitalize]{cleveref}
\crefname{section}{Sec.}{Secs.}
\Crefname{section}{Section}{Sections}
\Crefname{table}{Table}{Tables}
\crefname{table}{Tab.}{Tabs.}

\def\cvprPaperID{11793} 
\def\confName{CVPR}
\def\confYear{2022}

\begin{document}

\title{MixSyn: Learning Composition and Style for Multi-Source Image Synthesis}

\author{\.Ilke Demir\\
Intel\\
{\tt\small idemir@purdue.edu}
\and
Umur A. \c{C}ift\c{c}i\\
Binghamton University\\
{\tt\small uciftci@binghamton.edu}
}

\twocolumn[{%
\renewcommand\twocolumn[1][]{#1}%
\maketitle
\begin{center}
    \includegraphics[width=1\textwidth]{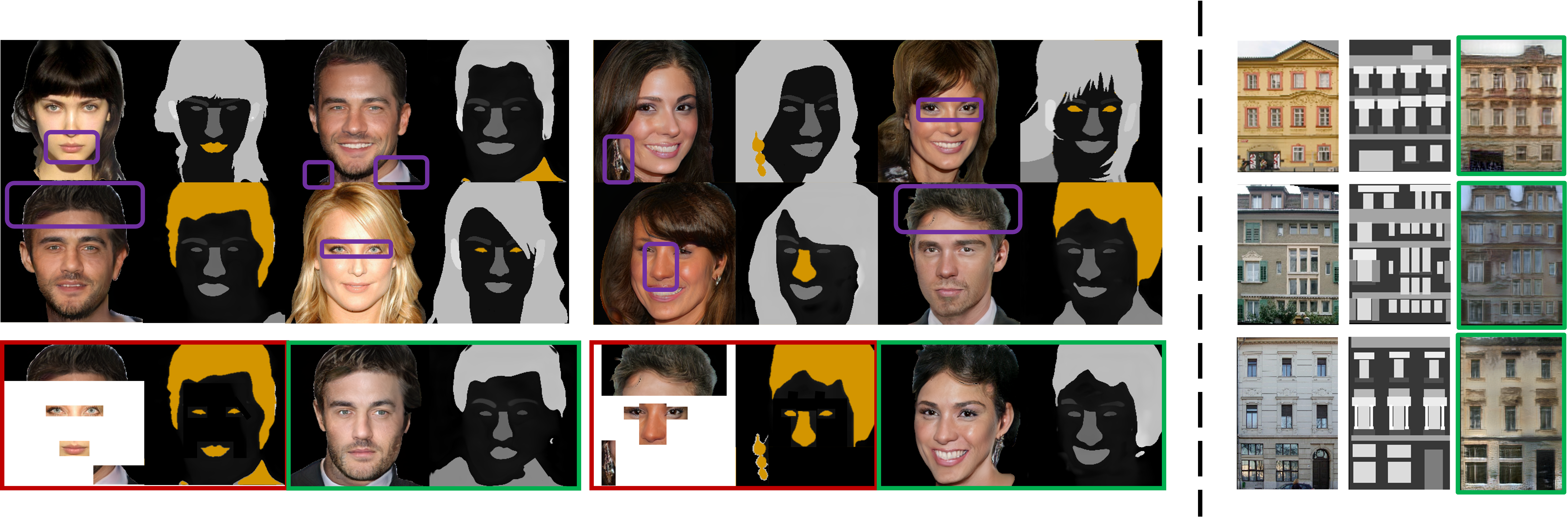}\\
    \textbf{Figure 1. }{MixSyn learns to generate semantic compositions and styles from multiple sources. Left - From mask (orange) and image (purple) regions, novel compositions and images (green) are generated. Naive copy-paste is shown in red boxes. \\ Right - Each facade (green) is generated from multiple source images for each region in the given mask. 
    }
    \label{fig:teaser}
\end{center}%
}]
\begin{abstract}\vspace{-0.2in}
Synthetic images created by generative models increase in quality and expressiveness as newer models utilize larger datasets and novel architectures. Although this photorealism is a positive side-effect from a creative standpoint, it becomes problematic when such generative models are used for impersonation without consent. Most of these approaches are built on the partial transfer between source and target pairs, or they generate completely new samples based on an ideal distribution, still resembling the closest real sample in the dataset. We propose MixSyn (read as ``mixin\' '') for learning novel fuzzy compositions from multiple sources and creating novel images as a mix of image regions corresponding to the compositions. MixSyn not only combines uncorrelated regions from multiple source masks into a coherent semantic composition, but also generates mask-aware high quality reconstructions of non-existing images. We compare MixSyn to state-of-the-art single-source sequential generation and collage generation approaches in terms of quality, diversity, realism, and expressive power; while also showcasing interactive synthesis, mix \& match, and edit propagation tasks, with no mask dependency.
\end{abstract}

\section{Introduction}
\label{sec:intro}
Image-based synthesis has been an interesting topic for decades in both computer vision and graphics. Recent generative approaches set this task forth as conditional generation~\cite{Bodla_2018_ECCV,Wang_2018_CVPR,pix2pixhd,lggan,ggwp,huh2020ganprojection,suzuki2019spatially}, image-to-image translation~\cite{stargan,huh2020ganprojection,Zhang_2018_ECCV,cafegan,disneyfaceswap,starganv2,chenDeepFaceDrawing2020,10.1007/978-3-030-58545-7_19}, or style encoding~\cite{softmasks,stylegan,adain,Collins_2020_CVPR}. The corner stone of these approaches has been learning the mapping between a source and a target image, for modeling specific styles, segments, or domains. Most of those approaches utilize semantic masks to conditionally generate realistic images~\cite{spade}, to represent diverse inter-domain images~\cite{starganv2}, and to replace the content or style of specific parts seamlessly~\cite{sean}. However, all of them \textit{operate on given masks} of \textit{source and target pairs}. Some enable sequentially modifying regions with multiple targets, but they need aligned segments with a constant mask.

Being able to incorporate style information through normalization parameters accelerated conditional image generation research, which yields higher quality results~\cite{adain} as the details are retained deeper in the network. However this constraint also increased the dependency on the input semantic masks (sometimes also called maps or compositions). Observing the state of the art in semantic image synthesis, three main limitations restrain the expressive power: (1) generation is restricted to source-target (pairwise) transfer of styles and regions, (2) semantic masks are mostly manually modified and they are neither novel, nor flexible, (3) uncorrelated and unaligned regions are compositionally not coherent. For (1),~\cite{sean} intakes per region style images, however they pursue pairwise processing per region. This is a serious limitation amongst most of the semantic image synthesis approaches as the interactions and contributions from multiple sources are dismissed. For (2),~\cite{spade} provides a UI for drawing semantic masks. However, the generation is based on the label encoding, and it is not possible to guide the generation with a specific image, sweeping the mask dependency under the hood. For (3),~\cite{ggwp} learns mapping and interpolation between masks, however our motivation to transcend pairwise manipulation to multi-source images synthesis poses a different challenge. Moreover, this source-target coupling enables impersonation by deepfakes, which raises serious ethical debates. 

To overcome these limitations, we jointly learn semantic compositions and styles from multiple images. We tackle this problem by learning to generate fuzzy semantic compositions from input masks and by learning to synthesize novel photorealistic images from these compositions, preserving the style of each input region. Although humans are comfortable editing existing semantic masks, manual assembly of novel masks from scratch is challenging due to (i) non-exact region boundaries, (ii) unassigned pixels, (iii) overlapping regions, and (iv) misalignment. MixSyn takes as input multiple \textit{unaligned} segments from several source images (i.e., eyes of A, mouth of B, and nose of C), and creates a new coherent image (i.e., a new face) based on the learned semantic maps (Fig.~\ref{fig:teaser}1). Our approach
\begin{itemize} 
\item learns to generate \textbf{coherent novel compositions}, reducing the dependency on semantic regions and increasing the quality; 
\item \textbf{couples structure and style generation} for image synthesis, flexing spatial constraints on the style generation by learned fuzzy masks; and
\item allows combining \textbf{multiple sources} into a photorealistic image, enabling style and structure blending, and disabling impersonation for face generation.
\end{itemize}
We employ two architectures for generating the composition (semantic) and the image (visual), encoding structures and styles of images separately per region. The structure generator (Fig.~\ref{fig:str}) learns feasible compositions from as-is, random, and real samples. The style generator (Fig.~\ref{fig:sty}) learns to generate realistic images using region-adaptive normalization layers with generated masks. The two generators are trained jointly in order to couple structure and style creation. We also introduce \textit{MS block} (Fig.~\ref{fig:sty}e) with optional normalization and resampling layers per module.

We demonstrate and compare our results to single-source sequential editing and collage-based synthesis approaches in terms of similarity, reconstruction, visual, and generative quality. We train and test MixSyn on several datasets in two domains: faces and buildings, with promising results for extension to others. We conduct ablation studies on our region classes and loss functions. Moreover, we implement several applications of MixSyn, such as edit propagation and combinatorial generative space exploration. The multi-source nature of MixSyn also prevents one-to-one impersonations in face domain, which is a positive step towards privacy concerns~\cite{probexists}, causing the shift to synthetic datasets~\cite{Wood_2021_ICCV}.

\section{Related Work}

\textbf{Patch-based Synthesis.} Traditional approaches provide semantically guided synthesis using patch similarity~\cite{10.1145/1531326.1531330}, graph cuts exploiting repetitions~\cite{graphcut}, and guided inverse modeling exploiting instances~\cite{gproc}. Their deep generative counterparts flex similarity and repetition coercion, so the synthesis can be much efficient~\cite{lee2016laplacian}, adaptive~\cite{Yang_2019_ICCV}, complex~\cite{tseng2020retrievegan}, yielding detailed results~\cite{Texler20}, due to simplistic part-based similarity~\cite{examplepatch} and contrastive~\cite{10.1007/978-3-030-58545-7_19} losses. Inspired by patch-based approaches, we propose a novel semantic image synthesis method where patches are replaced with fuzzy semantic regions, shifting our focus from patch selection to patch composition.

\textbf{Style Transfer.} Recently, popular image manipulation tasks emerge from applying the style of a source image to a target image by adaptive normalization~\cite{stylegan}, with explicit domain labels~\cite{stargan}, utilizing soft masks~\cite{softmasks}, transferring segment by segment~\cite{lggan}, for attribute editing~\cite{cafegan}, and in multiple domains~\cite{starganv2}. In particular for combining multiple sources,~\cite{suzuki2019spatially, Zhang_2018_ECCV} blend features in GAN layers of multiple reference images; however the spatial regions and blended features are provided manually.~\cite{michigan} conditions hair generation on multi-input; however masks are kept constant.~\cite{collage} can translate a collage image to a photorealistic image, but there is no semantic structure and the collage creation is a manual pre-processing step.
\stepcounter{figure}

        \begin{figure*}[ht]
        \begin{center}
           \includegraphics[width=0.9\linewidth]{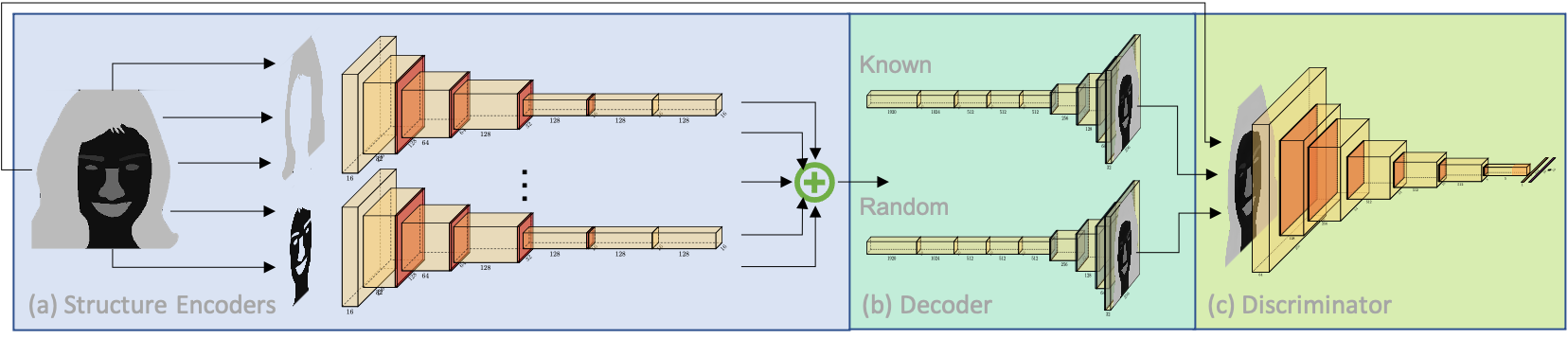}
        \end{center}
           \caption{\textbf{Structure Generator.} We train separate encoders for each region type (a), then the combined \textit{known} and \textit{random} structure codes are passed to the decoder (b). The generator and the discriminator (c) learns novel compositions.}
        \label{fig:str}
        \end{figure*}

\textbf{Conditional Normalization.} As semantic synthesis approaches and conditional GANs start to demand more accuracy and realism, supplying masks only as an input to the first layers did not suffice to preserve the contribution of regions as the network grows deeper. Later, the quality of results has been significantly enhanced by injecting style~\cite{adain} and structure~\cite{spade} information in the adaptive normalization layers.~\cite{sean} took it a step further and introduced region-adaptive normalization, which allows introducing per region styles. Building upon, we introduce MixSyn blocks (MS-block), a slight modification over~\cite{spade} with~\cite{sean} normalization, to broadcast styles per \textit{learned} regions.
        
\textbf{Semantic Editing.} Another semantic image manipulation direction is to modify or create some binary mask for inclusion/exclusion~\cite{huh2020ganprojection, sesame}, manipulate the underlying mask~\cite{10.1145/3306346.3323023, ggwp}, generate the mask only with label collections~\cite{cheng2020segvae}, replace foreground objects~\cite{Chen_2019_CVPR}, learn binary compositions~\cite{compgan}, use encoder-decoder networks to learn the blending~\cite{disneyfaceswap}, or infill with another image~\cite{deng2020referenceguided} to inpaint the manipulated parts. Although such approaches provide control over semantic labels, (1) generation is not controllable or guided by a certain image, (2) they mostly do inpainting instead of synthesis, and (3) there is no multi-source capability, i.e., all of them utilize source-target pairs. Meanwhile, other approaches push the image-to-image translation to mask-to-mask translation~\cite{maskgan}, sketch-to-sketch translation~\cite{Chen_2020_CVPR}, or scene graph editing~\cite{scenegraph}, where the new mask contains structure of the source and style of the target. Our approach is conceptually similar, but instead of user-defined masks, the mask is a learned composition of regions from multiple masks. 

\section{Multi-Source Composition Learning}

\begin{figure}[ht]
        \begin{center}
           \includegraphics[width=1\linewidth]{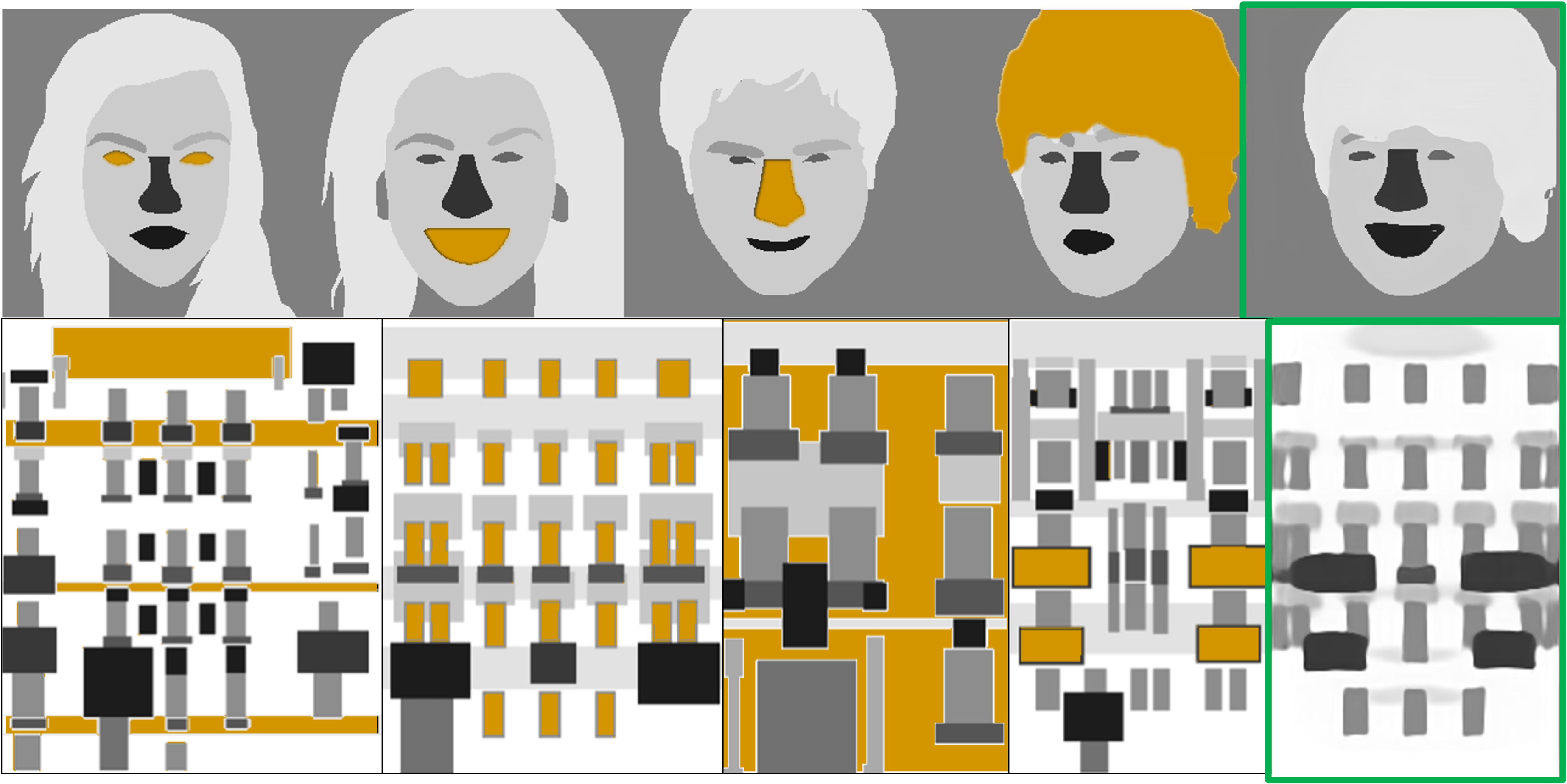}
        \end{center}
           \caption{\textbf{Compositions.} Orange regions ($r^i_j$) are used to generate random composition $M''$ (green) for faces (top) and buildings (bottom). More samples can be found in Supp. A.}
        \label{fig:str_img}
        \end{figure}

In order to learn coherent fuzzy compositions from multiple regions as in Fig.~\ref{fig:str_img}, first we define our compositions, then we describe our architecture with a multi-encoder, single decoder generator with a simple discriminator (Fig.~\ref{fig:str}).

\subsection{Compositions}
Let $r^a_i$ denote regions making up a source mask $M_a=\{r^a_i\}$, in a predetermined order for $i \leq N$, where $N$ is the number of all possible regions. $M$ corresponds to the list of all $S$ source masks $M=\{M_a, M_b, \dots, M_S\}$. We would like to assemble a composition $M'_*=\{r^a_0, r^b_1, \dots, r^c_N\}$ where each region $r^*_i$ comes from a source mask $M_*$ in $M$. It is important to note that a source mask can be selected multiple times for different regions ($a, b, \dots, S$ can repeat), however a region can be selected only once ($0, 1, \dots, N$ is unique). Masks have sharp boundaries between regions, whereas compositions combine fuzzy regions.
        
Needless to say, if all regions are selected from the same source mask (i.e., $\forall * = a $), we expect $M'_a=\cup_ie^a_i$ to represent $M_a$. We call this \textit{known} composition $M'_*$ for each $M_*$. In contrast, if each $r^*_i$ is selected from different $M_*$'s in the batch, we call it \textit{random} compositions $M''$ (Fig.~\ref{fig:str_img} and Supp. A). If a region does not exist in a composition, we set $e_{x}^*=\left[ 0\right]$. For the face domain, we select symmetric regions from the same source, e.g., left/ right eyes from one mask (see Supp. G for symmetry coupling), to keep random compositions consistent.  
    
             \begin{figure*}[ht]
        \begin{center}
           \includegraphics[width=0.9\linewidth]{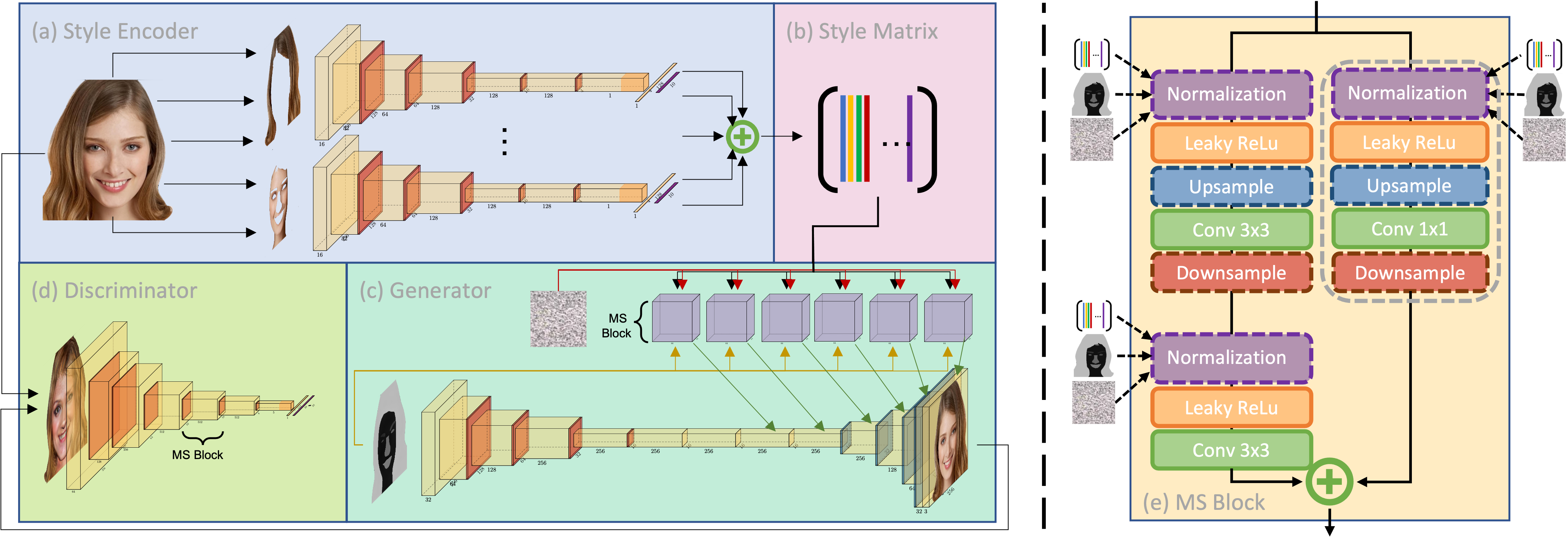}
        \end{center}
           \caption{\textbf{Style Generator.} We train $N$ encoders for each segment type (a), and create a style matrix (b). Style generator (c) translates the masks created by the structure generator into photorealistic images, using a region-adaptive normalization to broadcast the style matrix. \\ Figure. 4.e. \textbf{The MS Block.} Unit of processing for all of our networks is a res block with optional resampling and normalization layers.}
        \label{fig:sty}
        \end{figure*}

\subsection{Structure Generator}
There is no initial alignment between regions of random compositions, so it does not make sense to put many random regions into same composition in image space. However, we want to learn how they would transform and blend to create realistic compositions, thus we encode each $r^*_i$ with the specific region encoder $E_i(r^*_i)=e^*_i$, producing a $16\times16\times128$ structure code. We use separate encoders, so that the codes are disentangled and each region can be used interchangeably. Then, we combine structure code $e^*_i$'s into a composition code $c_*=\bigoplus_i e^*_i$  of size $16\times16\times128\times N$ and pass it to the decoder $C_*$. $C_*$ learns to decode $c_*$ into novel compositions $M_*$. Our structure generator forges soft borders (i.e., fuzzy compositions), which creates flexibility for our image generator to produce better results.  

The encoder-decoder structure constitutes our structure generator $G_M(M_*)=C_*(\bigoplus_{r^*_i \in M_*} E_i(r^*_i))$, which is trained with our discriminator $D_M$ to create coherent and realistic masks. The encoder, decoder, and discriminator models use MS blocks (Sec.~\ref{sec:moyd} and Fig.~\ref{fig:sty}e) and the layer details are documented in Fig.~\ref{fig:str} and Supp. B.

\subsection{Training Objectives}

During training, generator $G_M$ takes a mask $M_x$ and learns to create known compositions $M'_x$ and random compositions $M''$ with an adversarial loss. We also add $R_1$ regularization for training stability~\cite{r1reg}. The discriminator $D_M$ (Fig.~\ref{fig:str}b, Supp. B) aims to classify real masks $M$, generated known compositions $G_M(M')$ and generated random compositions $G_M(M'')$ (Fig.~\ref{fig:str}c, Supp. B). With a batch size of $\omega$, the discriminator processes $\omega$ reals and $2\omega$ fakes, thus we balance the contributions in the loss function. 
\begin{align}
    L_A=& log D_M(M_x) + \\\nonumber
     & \alpha log(1-D_M(G_M(M'_x))) + \\
     & (1-\alpha)log(1-D_M(G_M(M'')))\nonumber
\end{align}
For known compositions, we incorporate an L1 reconstruction loss $L_R= ||M_x-G_M(M'_x)||_1$, forming final objective:
\begin{align}
\min_{G_M} \max_{D_M} \lambda_A L_A + \lambda_R L_R
\end{align}

\section{Multi-Source Image Synthesis}
We continue with image generation from compositions, where each segment preserves its style. To help the reader, \textit{regions} in \textit{masks} are analogous to \textit{segments} in \textit{images}.

\subsection{Style Segments}
Let $k^a_i$ denote segments making up a source image $I_a=\{k^a_i\}$, with corresponding mask regions $M_a=\{r^a_i\}$. $I$ corresponds to the list of all $S$ source images $I=\{I_a, I_b, \dots, I_S\}$. We would like to assemble an image $I'_*=\{k^a_0, k^b_1, \dots, k^c_N\}$ where each region $r^*_i$ corresponding to the segment $k^*_i$ comes from a mask $M_*$ in $M$. The concept can be observed in red copy-paste segments in Fig.~\ref{fig:teaser}1.

We utilize three types of segments: (1) $\{k^a_i\}$ from $\{r^a_i\}$ in the initial mask $M$, (2) $\{k'^a_i\}$ from $\{r'^a_i\}$ in the generated known composition $G_M(M'_a)$, and (3) $\{k''^*_i\}$ from $\{r''^*_i\}$ in the generated random composition $G_M(M'')$. We expect $\cup_i E_i(k^a_i)$ to represent $I_a$, and $\cup_iE_i(k'^a_i)$ to approximate $I_a$. While these two segment generations ensure learning plausible photorealistic images from compositions, the last one ($\cup_iE_i(k''^a_i)$) is the actual novelty that brings out the style blending from multiple source images. This is also depicted as the main application in Fig.~\ref{fig:creation}.


Similar to the structure generator, we encode each $k^*_i$ with the specific segment encoder $E_i(k^*_i)=e^*_i$, producing a $\delta$-length style code (Fig.~\ref{fig:sty}b). Then we combine the style codes to construct a style matrix $\Delta_*=\bigoplus_i e^*_i$ of size $\delta\times N$. Encoder layers are listed in Fig.~\ref{fig:sty}a and in Supp. B.

\subsection{Image Generator}
We use a full generator with adaptive normalization layers for image synthesis $G_I(M_*,\Delta_*)=I_*$ (Fig.~\ref{fig:sty}c), which is trained with our image discriminator $D_I$ (Fig.~\ref{fig:sty}d) to create realistic images. Supp. B delineates all architectures.

\subsubsection{MS Block}\label{sec:moyd}
To selectively include normalization and sampling layers throughout our architecture, we introduce our minimum computation unit: MS block (Fig.~\ref{fig:sty}e). MS is a configurable res block with a shortcut, with optional downsampling (red layers in Fig.~\ref{fig:str}), upsampling (blue layers in Fig.~\ref{fig:sty}c), and normalization (purple boxes in Fig.~\ref{fig:sty}c) layers. Samplings are done with bilinear interpolation and average pooling. For encoders and structure decoder, instance normalization is enabled in MS block. For broadcasting styles per learned regions, we use region-adaptive normalization~\cite{sean} with corresponding masks and style matrices. Layer order in MS block follows pre-activation residual units in~\cite{postres,starganv2}.

\subsubsection{Training Objectives}

\textbf{Adversarial Loss.} Our image generator $G_I$ intakes source images $I$ and compositions $M_x$, $G_M(M'_x)$ and $G_M(M'')$, outputting known $G_I(I_x,M_x)$, approximated $G_I(I_x, G_M(M'_x))$, and random images $G_I(I_*, G_M(M''))$. The discriminator $D_I$ (Fig.~\ref{fig:sty}d and Supp. B) classifies these images as real or fake using loss~\ref{eqn:adv}, balancing contributions of real and three subsets of fake images. We add $R_1$ regularization for training stability~\cite{r1reg}.
\begin{align}\label{eqn:adv}
    L_A=& \beta \log D_I(I_x) + \\\nonumber
     & (1-\beta)\left[ \eta (\log(1-D_I(G_I(I_x,M_x))) + \right. \\\nonumber
     & \ \ \ \ \ \ \ \ \ \ \ \ \ \ \ \ \ \log(1-D_I(G_I(I_x, G_M(M'_x))))) + \\\nonumber
     & \ \ \ \ \ \ \ \ \ \ \ \ \ \left. (1-\eta) \log(1-D_I(G_I(I_*, G_M(M'')))) \right] \nonumber
\end{align}
Note that, initial ${r^*_i}$s from different $M_*$s that are combined in $M''$ are stored in order to evaluate the corresponding ${k^*_i}$s in $I_*$. Although the region-adaptive normalization layers need $\Delta$, we push the extraction of the style matrix per composition, to better fill approximate masks. 

\textbf{Style Loss.} We add loss~\ref{eqn:sty} based on style matrix $\Delta_*=\bigoplus_i E_i(k^*_i)$ to ensure that the style is preserved for segments of approximated and random images that undergo some transformation. 
\begin{align}\label{eqn:sty}
    L_S= \frac{1}{2N}  (& ||\bigoplus_i e^x_i - \bigoplus_i E_i(G_I(I_x, G_M(M'_x)))|| +\\\nonumber
    & ||\bigoplus_i e^*_i - \bigoplus_i E_i(G_I(I_*, G_M(M'')))||) \nonumber
\end{align}

\textbf{Reconstruction Loss.} Similar to the structure generator,  we incorporate a reconstruction loss for the known and approximated images, as they originate from the same image.
\begin{align}
L_R= \frac{1}{2}( & ||I_x-G_I(I_x, M_x)||_1 +\\\nonumber
&||I_x-G_I(I_x, G_M(M'_x))||_1 )
\end{align}
Formulating a piecewise continuous local reconstruction loss (like~\cite{ggwp}) for random images is left for future work.

Overall, our training can be formulated as below, with the corresponding hyperparameters for each loss term.
\begin{equation}
\min_{G_I,E} \max_{D_I} \lambda_A L_A + \lambda_S L_S + \lambda_R L_R
\end{equation}

\section{Results}
We set 0.0001 and 0.0003 for the learning rates of $G_M$, $G_I$ and $D_M$, $D_I$, using ADAM~\cite{adam} with $\beta_1=0$ and $\beta_2=0.999$ with a decay of 0.0001. Similar to other normalization approaches~\cite{sean, spade}, we apply Spectral Norm~\cite{spec} to generators and discriminators. We use instance and region-adaptive normalization for specified layers (see Supp. B). Experiments are done on an NVIDIA RTX 2080 with 4 GPUs/
We use 30000 images in CelebAMask-HQ~\cite{maskgan} for most of the experiments in face domain, Helen~\cite{helen} for cross-dataset evaluation, and CMP Facade dataset~\cite{CMPDatabase} for results in architecture domain. We use SSIM~\cite{ssim}, RMSE~\cite{psnr}, PSNR~\cite{psnr}, and FID~\cite{fid} scores for quantitative evaluations and comparisons.

\subsection{Evaluation}

\begin{figure}[h]
\begin{center}
   \includegraphics[width=1\linewidth]{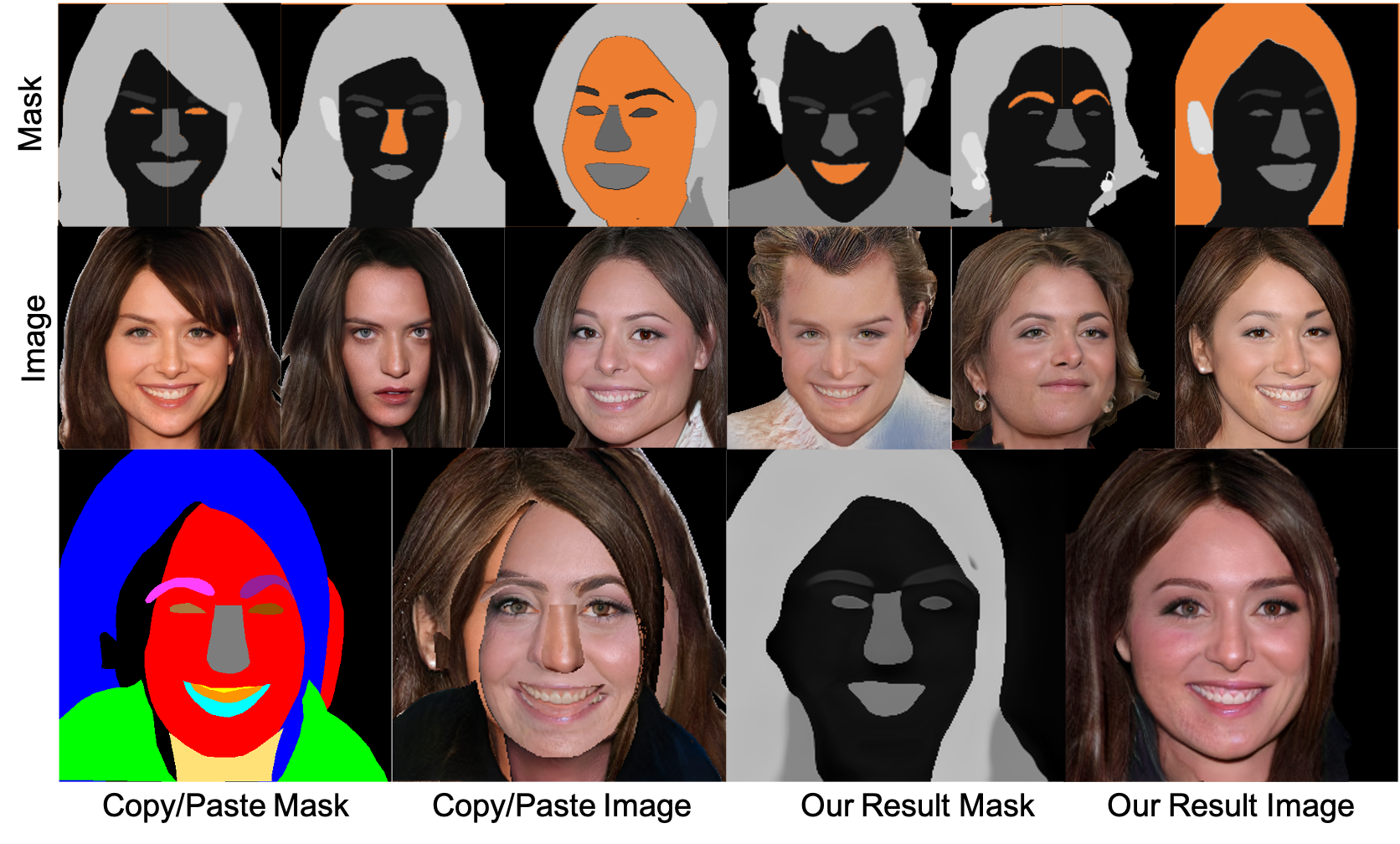}
\end{center}
   \caption{\textbf{Multi-Source Synthesis.} MixSyn creates a composition and image (right) from six regions (orange) of six segments (mid).
   }
\label{fig:creation}
\end{figure}

Fig.~\ref{fig:creation} demonstrates our main purpose. If selected regions (orange) are to be naively copy-pasted, bottom left mask-image pair is obtained, which is not desirable. In contrast, our approach is able to combine six segments from six images into a coherent composition and image (bottom right). Fig.~\ref{fig:rezilimage} shows examples of generated composition and image pairs for buildings with 4+ different sources, whereas Fig.~\ref{fig:ultimo} and 1 demonstrate results (green) for faces, as a seamless combination of purple segments from 3, 4, and 5 source images. 
Note that purple boxes are not exact, they only mark the region which is actually represented with the orange masks (e.g., box on a head represents \textit{all} hair in orange hair region). The copy-paste versions are only demonstrated as examples, they are not used in MixSyn.

We evaluate our method quantitatively in Tab.~\ref{tab:scorecomp} with different image similarity metrics applied per region. We also document region-based scores in Supp. C-E, revealing that our bottleneck is to learn hair styles (highest FID). MixSyn realistically generates frequent segments (eyes, nose, mouth) with high SSIM and PSNR, however similarity scores of rare ones (hat, glasses) are much lower. 

   \begin{table}[h]
\begin{center}
\begin{tabular}{c|c|c|c|c}
\hline
Method & SSIM &  RMSE &  PSNR &  FID \\
\hline
Pix2PixHD~\cite{pix2pixhd} & 0.68 &  0.15  &   17.14  &  23.69\\
SPADE~\cite{spade}& 0.63& 0.21   &  14.30 &   22.43\\
SEAN~\cite{sean} & 0.7& \textbf{0.12} &    18.74 &   17.66\\
MixSyn  Str & 0.97 & 1.15 & 33.06 &  18.13\\
MixSyn & \textbf{0.95} & 1.89 & \textbf{31.32} &  \textbf{14.41}\\
\hline
MixSyn Str (H) & 0.98 & 0.92 & 36.00 & NA \\
MixSyn (H) & 0.96 & 1.46 & 32.13 & NA\\
\hline
\end{tabular}
\end{center}
\caption{\textbf{Reconstruction Scores} on CelebAMask-HQ and Helen (H) datasets. Non-MixSyn scores are taken from~\cite{sean}. \label{tab:scorecomp}}
\end{table}

Finally, we perform a cross-dataset evaluation and test MixSyn trained on CelebAMask-HQ on Helen (Tab.~\ref{tab:scorecomp} (H)). High similarity indicates that MixSyn is generalizable to create multi-source faces from other datasets. Relatively lower RMSE signals that we indeed create novel (fuzzy) masks \textit{with inexact reconstructions} where multiple regions adapt and blend. Supp. E declares all cross-dataset scores.


\subsection{Comparison}

As MixSyn is the first of its kind, we compare our results to single-source~\cite{spade,starganv2}, sequential multi-source~\cite{ggwp,sean}, and collage-based~\cite{collage} approaches. These approaches (i) cannot generate from multiple sources simultaneously, (ii) depend on given/modified mask, (iii) cannot compose novel masks, (iv) do not learn BOTH structure and style end-to-end, and (v) cannot generate from partial or fuzzy masks. 

    \begin{figure}[h]
\begin{center}
   \includegraphics[width=1\linewidth]{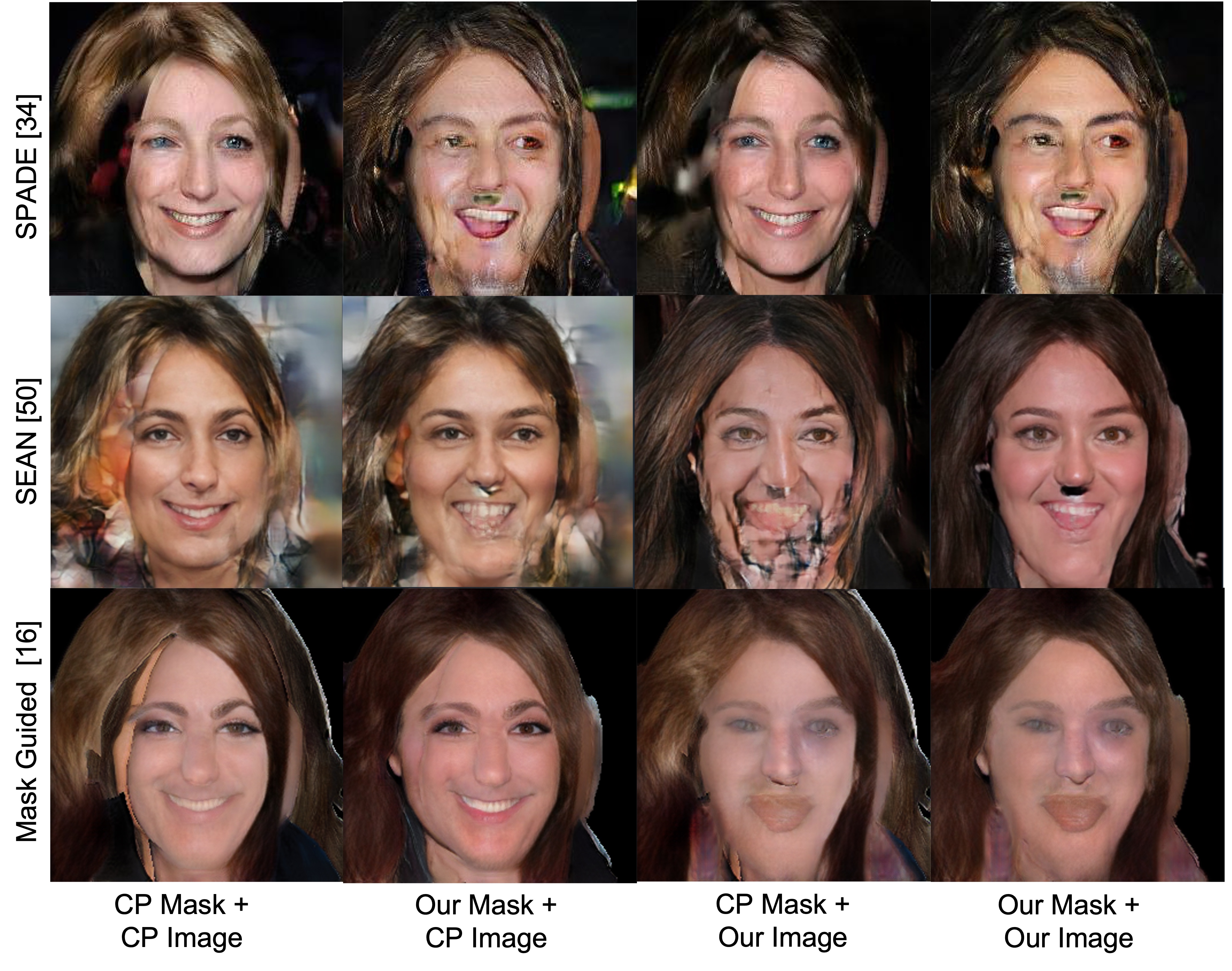}
\end{center}
   \caption{\textbf{Comparison}. Output pairs in Fig.~\ref{fig:creation} bottom, are fed to \cite{spade, sean, ggwp}. None can generate from multi-source, create non-existing masks, or output realistic results similar to originals. }
\label{fig:comp}
\end{figure}

We start with justifying these claims. As per (i-ii), we feed four combinations of (copy-paste/our) x (mask/image) pairs in Fig.~\ref{fig:creation} as alternative inputs to SPADE~\cite{spade}, SEAN~\cite{sean} and Mask Guided CGAN~\cite{ggwp}. Although results improve from copy-paste masks (Fig~\ref{fig:comp}, col. 1 \& 3) to our generated masks (2 \& 4), quality of their results are not close to ours (Fig~\ref{fig:creation}), supporting (ii-iii). We also investigate how others reconstruct our result image with our mask (col. 4). Because (iv-v) above, they simply cannot. 

\begin{figure}[h]
\begin{center}
   \includegraphics[width=1\linewidth]{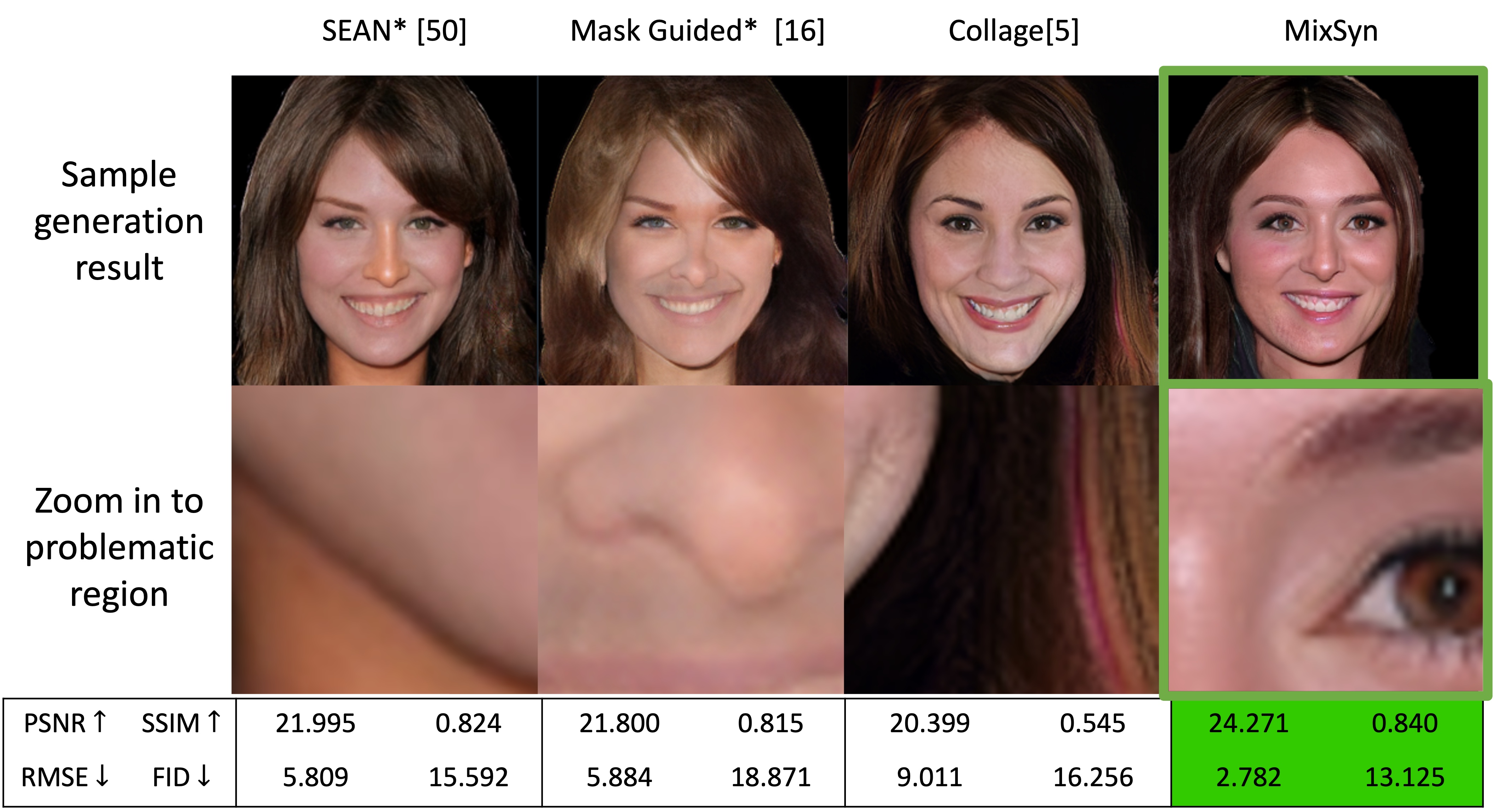}
\end{center}
   \caption{\textbf{Sequential*/Collage Comparison}. Component transfer causes artifacts on nose and neck for~\cite{ggwp}, and a ghost mustache for~\cite{sean}. Collage synthesis~\cite{collage} creates color artifacts and empty areas. See Supp. F for detailed image scores.}
\label{fig:comp2}
\end{figure}

In Fig.~\ref{fig:comp2}, we select a base mask (top left in Fig.~\ref{fig:creation}) because of (ii-iii), and swap segments following the sequential (hence (i)) component transfer applications of~\cite{sean, ggwp}, shown in the first two columns. Despite looking better than col. 1-4 in Fig.~\ref{fig:comp}, it is akin to a blended copy-paste, which creates the zoomed-in artifacts (e.g., different neck and nose colors, and shadow mustache), because they are not jointly composing new masks and generating new images as MixSyn does (iv-v). For the third column, we use the copy-paste image as the collage for~\cite{collage} input. Although there are less visual artifacts compared to the other two, there are empty areas and inconsistent hair style within the segment. The difference in similarity scores also proves that styles per regions are not as-well preserved as ours, questioning the realism over fidelity of~\cite{collage}. 


Quantitative comparison of MixSyn also supports and generalizes these claims. In Table~\ref{tab:scorecomp}, we list SSIM, RMSE, PSNR, and FID scores of Pix2PixHD~\cite{pix2pixhd}, SPADE~\cite{spade}, SEAN~\cite{sean}, 
our structure generator, and overall MixSyn architecture on CelebAMask-HQ dataset~\cite{maskgan}. Although our reconstruction is not as exact (worse RMSE), our generator network is better (better FID). We note that from our compositions to our images, similarity decreases (better SSIM and PSNR for MixSyn Str) as expected, but our style generator exploits novel compositions and achieves a better FID. We list same metrics for the example in Fig.~\ref{fig:comp2}, which are also better than SOTA. Please also check detailed reconstruction scores (Supp. C.), region similarity scores for \textit{random} images (Supp. D), and rest of Fig.~\ref{fig:comp} scores in Supp.~F.


\subsection{Experiments}
    
\begin{figure}[ht]
        \begin{center}
           \includegraphics[width=1\linewidth]{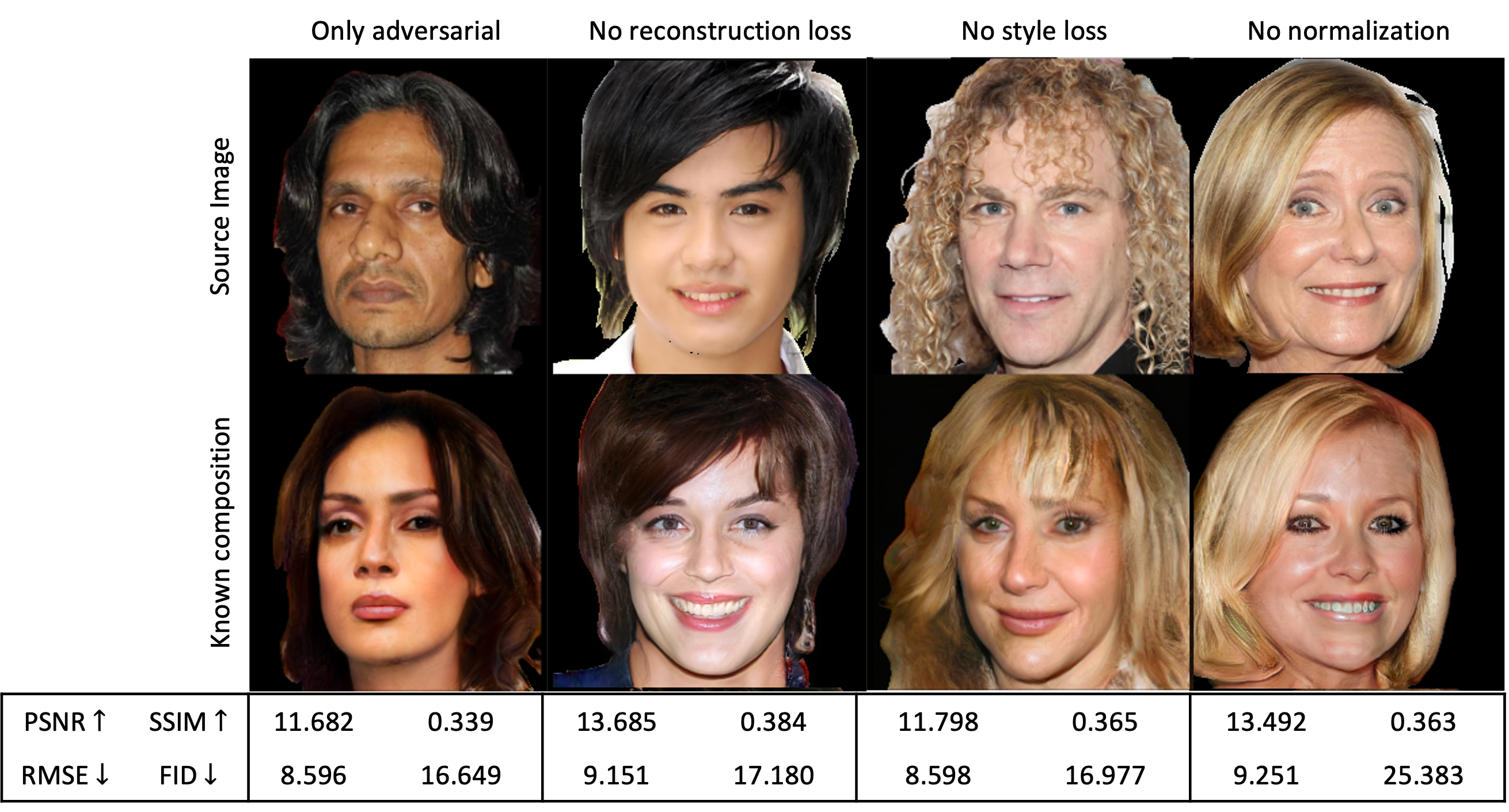}
        \end{center}
           \caption{\textbf{Ablation Study.} Samples of source and known image with different losses, followed by dataset scores for each setting.}
        \label{fig:abla}
    \end{figure}

Fig.~\ref{fig:abla} demonstrates and documents the contribution of each loss function. With only adversarial loss, we generate some humans fitting to compositions, but neither color, nor style, and not even the domain is preserved. Without reconstruction loss, we are able to mimic the style, but the colors are off. Without style loss, we lose patterns of each region, e.g., curly hair is ironed, even though they are from the same region. Finally, without region adaptive normalization, style of small regions are dominated (e.g., eyes). The dataset scores below are computed similar to Tab~\ref{tab:scorecomp}, but on the results generated with the specific loss functions.
    
\begin{figure}[h]
        \begin{center}
           \includegraphics[width=1\linewidth]{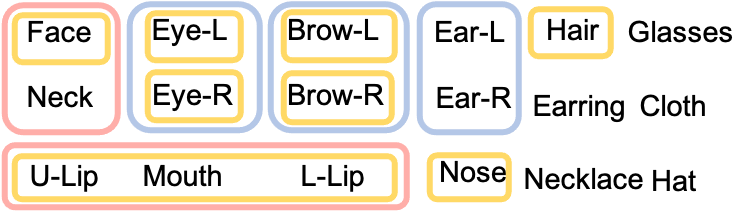}
        \end{center}
          \caption{\textbf{Region Types}. Starting from MaskGAN~\cite{maskgan} types, we create meta-types (pink), and couple symmetries for random generation (blue). We also experiment with compact types (yellow).}
        \label{fig:classes}
        \end{figure}

Another key construct is the selection of region types. 18 base types and their hierarchy are mostly known~\cite{maskgan} (Fig.~\ref{fig:classes}). However, for our problem, we experiment (i) without symmetry coupling for random compositions, and (ii) with 5 types only, before we converge on (iii) meta types.

\begin{figure}[h]
        \begin{center}
           \includegraphics[width=1\linewidth]{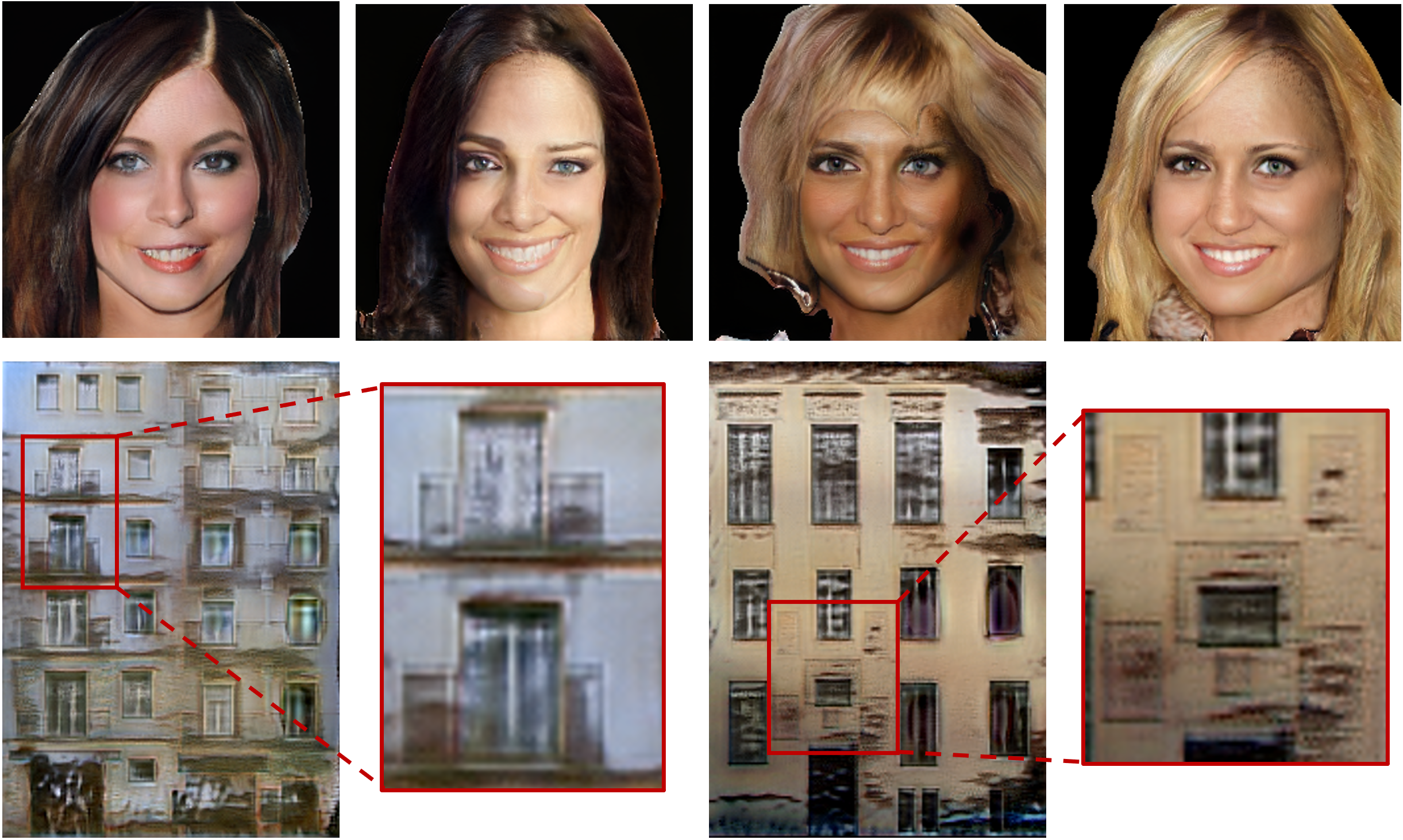}
        \end{center}
           \caption{\textbf{Without Symmetry Coupling}, random generation creates faces akin to uncanny valley, with unmatched colors and brows (top); and buildings with phantoms (bottom).}
        \label{fig:sym}
        \end{figure}

Early signs for detecting fakery were broken symmetries, such as mismatched eyes and brows. To enforce learning correlation of symmetric regions, we couple left-right indices in \textit{random} compositions (Fig.~\ref{fig:classes}, blue) for faces. Similarly for buildings, we coupled windows, cornices, and sills together for preserving patterns in random compositions. We experimentally validate that it is better than putting them into same channel. Fig.~\ref{fig:sym} shows results without symmetry coupling. Although they look realistic at a first glance, different eye colors, gaze directions, and eyebrow styles give away their synthetic nature, shifting faces to the uncanny valley. When those regions are selected randomly without following the same pattern in buildings, less dominant classes such as cornices and sills start to appear as phantoms on the buildings, as shown in the zoom ins. We also tried compact subtypes of face regions (6 yellow in Fig.~\ref{fig:classes}). We expected style generator to fill in rare types such as necklace, hat, etc. Instead, structure network merged them to existing types, creating interesting compositions (Supp. H). 

To decrease training time, increase accuracy, and fit encoders in the memory, we introduce meta-types by grouping. We intuit that finer granularity regions are needed for better style transfer, but not for synthesis. In other words, preserving the mouth as a whole is easier than learning the combination of lips, since we already create novel masks. 15 final meta-types are listed in Fig.~\ref{fig:classes} in pink. 

\section{Applications}

\subsection{Combinatorial Diversity}

\begin{figure}[h]
        \begin{center}
           \includegraphics[width=1\linewidth]{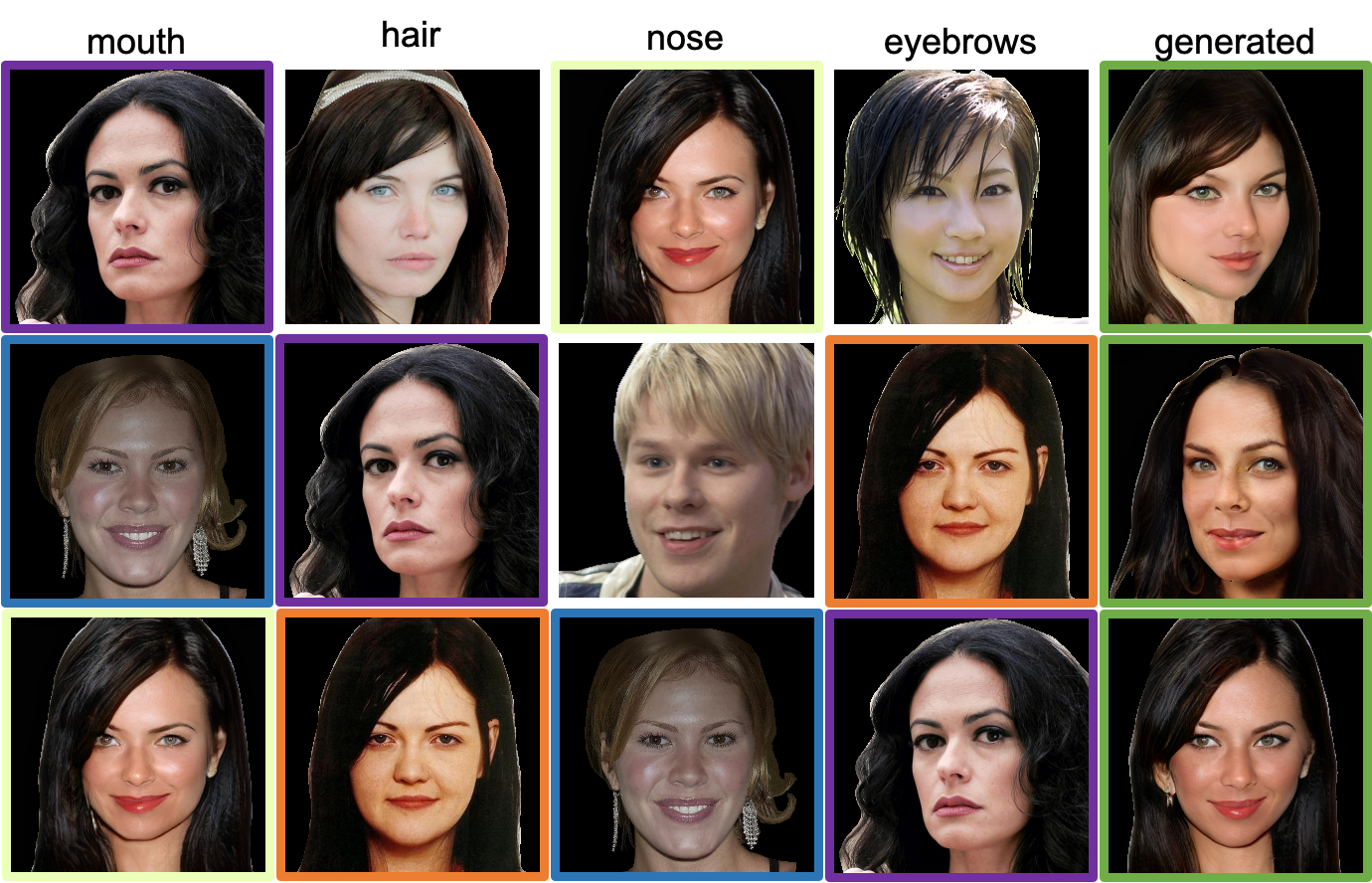}
        \end{center}
           \caption{\textbf{Design Space}. Using varied regions from same set of images (color-coded), design space grows exponentially (green). }
        \label{fig:combin}
        \end{figure}

Each row in Fig.~\ref{fig:combin} demonstrates combinations of different regions (mouth, hair, etc.) from similar sets of reference images (color-coded pairs), to create visually varying faces (green). As we can create an exponentially diverse set of combinations, we claim that such a combinatorial design space enables interactive editing systems, simulations with synthetic collections, and data augmentation for DNNs.

\subsection{Edit Propagation}

\begin{figure}[h]
        \begin{center}
           \includegraphics[width=1\linewidth]{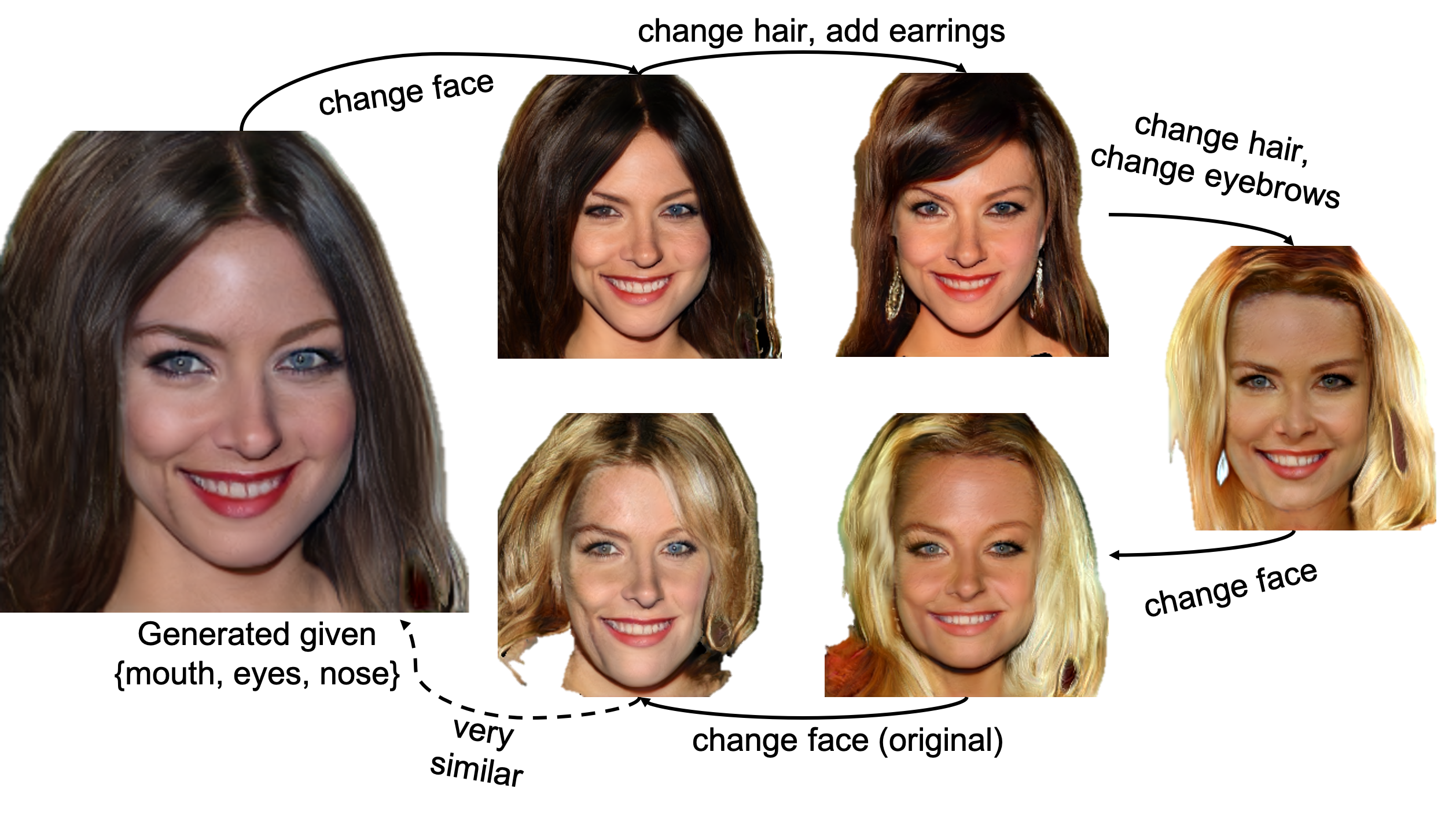}
        \end{center}
           \caption{\textbf{Perpetual Edits.} After first face (left) is created, each segment is replaced by others from different faces. Bottom left face is very similar to the first, because original face is swapped.}
        \label{fig:edit}
        \end{figure}

\begin{figure}[ht]
        \begin{center}
           \includegraphics[width=1\linewidth]{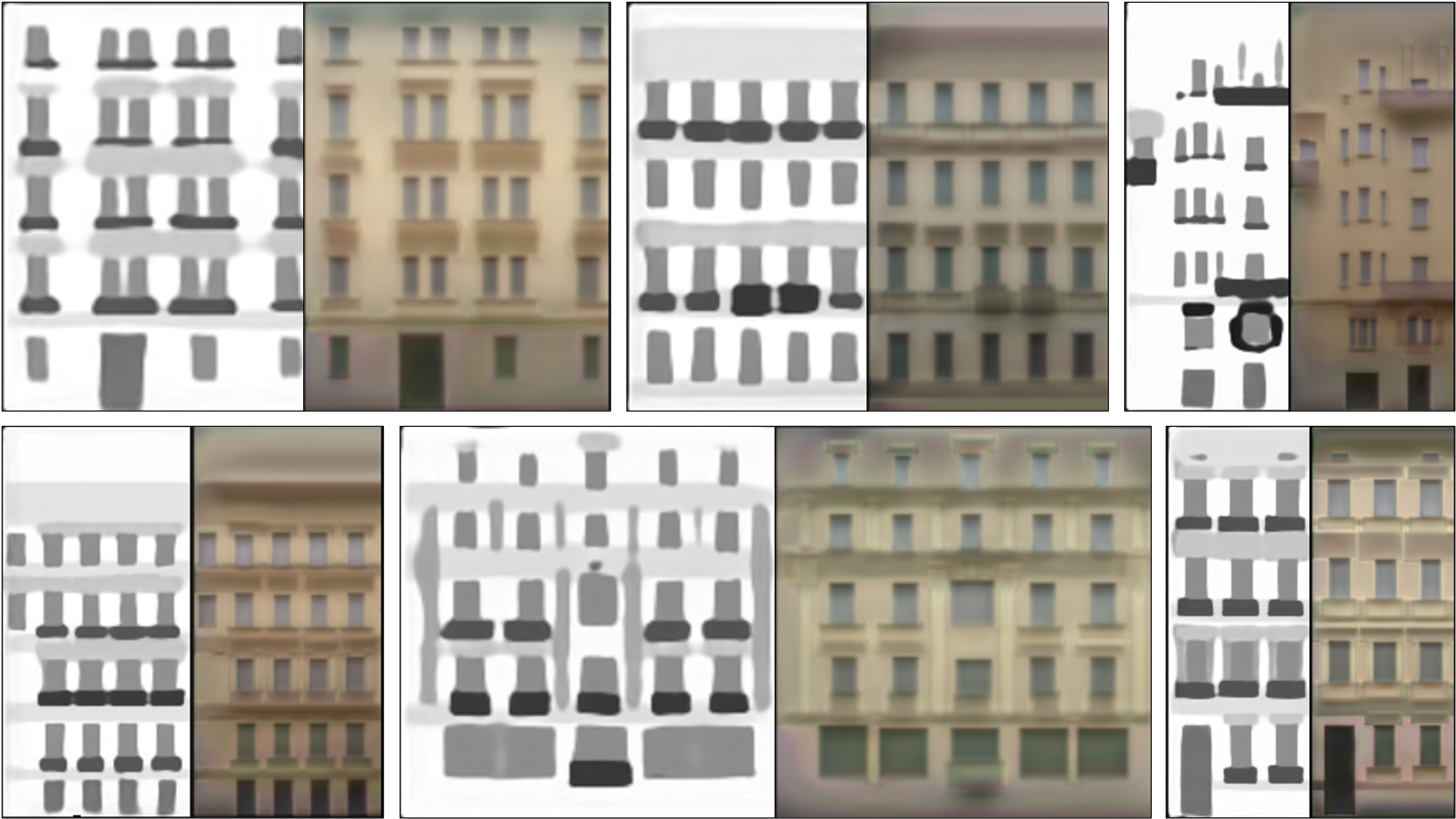}
        \end{center}
           \caption{\textbf{Synthetic Buildings.} Sample composition and image pairs generated in architecture domain, each region is selected from a different source building.}
        \label{fig:rezilimage}
        \end{figure}

\begin{figure*}[ht]
        \begin{center}
           \includegraphics[width=1\linewidth]{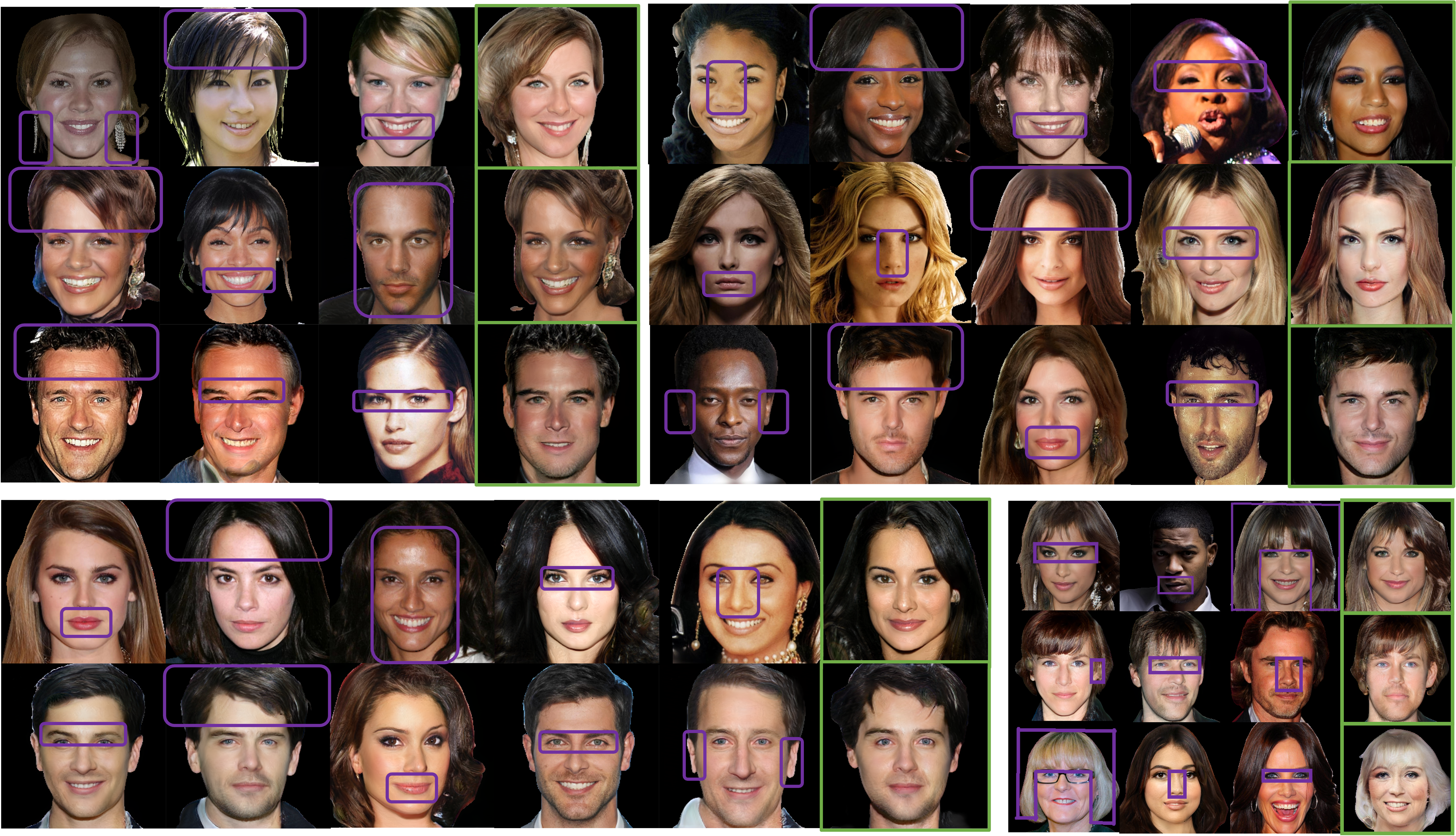}
        \end{center}
           \caption{\textbf{Additional Results.} Purple-highlighted segments in the first 3, 4, or 5 columns are used to synthesize new images (green). 
           }
        \label{fig:ultimo}
        \end{figure*}

In Fig.~\ref{fig:edit}, we start by generating an image given a set of segments, such as $\{$mouth, nose, eye$_l$, eye$_r\}$. Then, we change one or multiple segments with other known or suggested ones. Observe that other segments are structurally and stylistically preserved at each step, while the specified segments are changed according to an unseen reference. In the last step, the face in the first image is given as reference to change the face, creating a very similar face as the region features are preserved throughout the edits.

 \section{Discussion \& Limitations}

While generating an image, not all structure codes are needed, e.g., there is no cloth in input regions of Fig.~\ref{fig:creation}, so $e_{cloth}^*=\left[ 0\right]$. As there is some fuzziness between hair/face regions in third and last columns, both structure and style generators can recover it in a random composition, and place cloth region in mask, and cloth segment in image. On the other hand, some random combinations cause edge cases naturally, such as a random combination of face region from an image with hair and hair region on the sides from a bald person (Supp. H). An interactive editing system can aid in eliminating such random combinations.
As a synthesis approach, MixSyn cannot be used for analysis of data for harming populations. It can only create novel samples based on the training datasets or editing operations. Furthermore, it is almost guaranteed that the synthesized image is either a combination of parts from multiple images, or the same image; thus, it cannot be used for retargetting/reanimation/impersonation of existing people, causing misinformation. 
As a positive impact, we hope that our approach can spearhead anonymization efforts for sensitive data, when only a part of an image is needed.

\section{Conclusion and Future Work}
We introduce \textit{mixed synthesis (MixSyn)} for generating photorealistic images from multiple sources by learning semantic compositions and styles simultaneously. We train structure and style generators end-to-end, while preserving details by adaptive normalization on learned regions. We introduce a flexible MS block as the unit of processing for semantic synthesis. We demonstrate our results on three datasets and two domains, report our FID, SSIM, RMSE, and PSNR scores, qualitatively and quantitatively compare to prior work, and propose novel applications.

We observe that controlled synthesis with multiple images brings a new dimension to expressive creation. Our approach helps create non-existing avatars or architectures. It enables partial manipulation, region transfer, and combinatorial design without mask editing. Anonymization and de-identification are also facilitated by MixSyn. Finally, with the proliferation of adaptive normalization, multi-source synthesis will bloom, foreseeing MixSyn as a pioneer.

{\small
\bibliographystyle{ieee_fullname}
\bibliography{egbib}
}

\end{document}